\documentclass[10pt,twocolumn,letterpaper]{article}

\usepackage{cvpr}              

\usepackage{xcolor}
\definecolor{sgreen}{rgb}{0.0, 0.5, 0.0}

\definecolor{cvprblue}{rgb}{0.21,0.49,0.74}
\usepackage[pagebackref,breaklinks,colorlinks,allcolors=cvprblue]{hyperref}

\def\paperID{7320} 
\def\confName{CVPR}
\def\confYear{2025}

\title{Learning Velocity and Acceleration: Self-Supervised Motion Consistency for
Pedestrian Trajectory Prediction}

\author{Yizhou Huang\textsuperscript{1}, 
Yihua Cheng\textsuperscript{2}, Kezhi Wang\textsuperscript{1}\thanks{Corresponding author}\\
\textsuperscript{1}Brunel University of London, \textsuperscript{2}University of Birmingham\\
{\tt\small \{byyh009, Kezhi.Wang\}@brunel.ac.uk, y.cheng.2@bham.ac.uk}\\ 
}

\begin{document}
\maketitle
\begin{abstract}
Understanding human motion is crucial for accurate pedestrian trajectory prediction. Conventional methods typically rely on supervised learning, where ground-truth labels are directly optimized against predicted trajectories. This amplifies the limitations caused by long-tailed data distributions, making it difficult for the model to capture abnormal behaviors. In this work, we propose a self-supervised pedestrian trajectory prediction framework that explicitly models position, velocity, and acceleration. We leverage velocity and acceleration information to enhance position prediction through feature injection and a self-supervised motion consistency mechanism.  Our model hierarchically injects velocity features into the position stream. Acceleration features are injected into the velocity stream. This enables the model to predict position, velocity, and acceleration jointly. From the predicted position, we compute corresponding pseudo velocity and acceleration, allowing the model to learn from data-generated pseudo labels and thus achieve self-supervised learning. We further design a motion consistency evaluation strategy grounded in physical principles; it selects the most reasonable predicted motion trend by comparing it with historical dynamics and uses this trend to guide and constrain trajectory generation. We conduct experiments on the ETH-UCY and Stanford Drone datasets, demonstrating that our method achieves state-of-the-art performance on both datasets.
\end{abstract}    
\section{Introduction}
\label{sec:intro}
\begin{figure}[ht]
\begin{center}
\centerline{\includegraphics[width=\columnwidth]{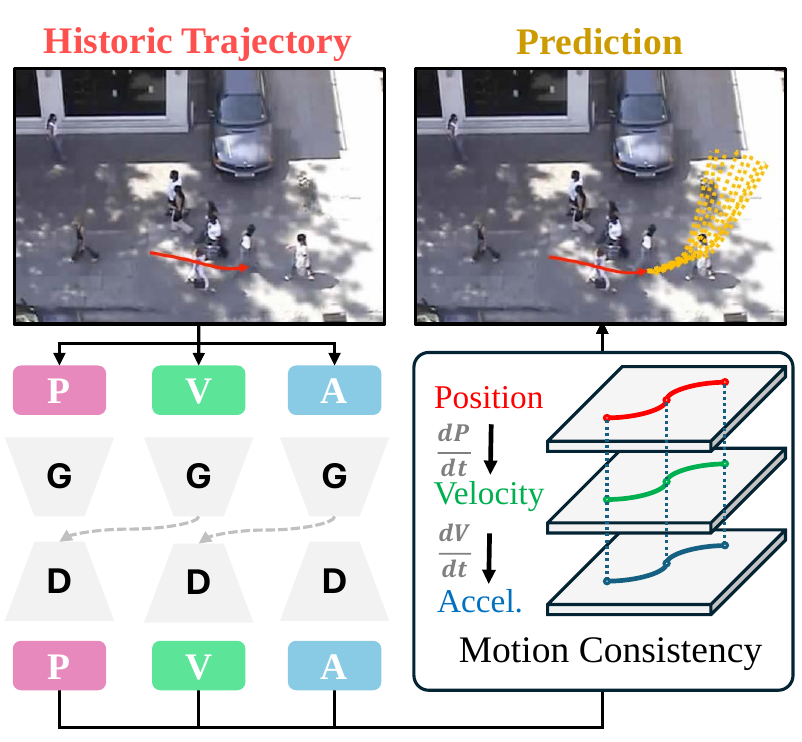}}
\caption{Our method takes historical pedestrian trajectories as inputs. From the historical trajectory, we extract position, velocity, and acceleration (accel.) information, which are then fed into a three-stream network. In the figure, the letter $d$ represents the amount of change. The network hierarchically fuses features across the three streams and enforces motion consistency among position, velocity, and accel predictions during training. The output consists of multiple possible future position trajectories.}
\label{Figure 1}
\end{center}
\end{figure}
Pedestrian trajectory prediction involves forecasting future
trajectories based on historical position sequences of pedestrians. This task is crucial in various real-world applications, including the management of intelligent transportation
systems \cite{li2024integrated}, enhancing the reliability of autonomous driving \cite{Cheng_2024_CVPR,huang2025trajectory}, guiding public safety efforts \cite{jafari2024pedestrians} and improving the understanding of human-robot interactions \cite{thumm2024human}. In these contexts, the accuracy of movement prediction is of paramount importance.

Conventional pedestrian trajectory prediction methods \cite{yuan2021agentformer,yue2022human,bae2023set,bae2023eigentrajectory} typically rely on the supervised learning paradigm \cite{sarker2021machine} as predicted trajectories are optimized by computing losses against ground-truth labels. This makes supervised learning inherently constrained by the long-tail distribution in trajectory prediction tasks. As a result, the model struggles to accurately capture abnormal or highly dynamic behaviors, such as sudden turns, stops, or evasive maneuvers. This limitation arises because supervised learning typically optimizes for overall loss, which is dominated by frequent behaviors. Consequently, the model tends to prioritize the prediction accuracy of majority patterns, while the contributions of rare behaviors are underrepresented, leading to potential overfitting to high-frequency trajectories and poor generalization in edge cases. This drew our attention and led to two key reflections:

1.) Can a self-supervised prediction framework be used in trajectory forecasting to mitigate overfitting?

2.) How can self-supervised learning use pseudo labels to guide the output of predicted trajectories?

In this paper, we propose a novel pedestrian trajectory prediction framework that explicitly models position, velocity, and acceleration\footnote{we omit the explicit mention of sequences in the remainder of this paper}. We generate three prediction outputs—position, velocity, and acceleration. Through a novel motion consistency mechanism, we evaluate the alignment between the predicted motion trend (derived from velocity and acceleration) and the observed motion trend. This consistency is then used to guide the refinement of the predicted position. As a result, the optimization of position is achieved automatically through the predicted velocity and acceleration, which serve as pseudo-labels. This constitutes a self-supervised learning approach.

Specifically, our method learns to predict velocity and acceleration while leveraging these predictions to impose motion consistency constraints on position prediction.
We compute the average velocity and acceleration from position and observed time steps, then use them as additional inputs, as demonstrated in Figure 1.
Our method takes as input a sequence of historical positions, velocities, and accelerations and outputs multiple potential future positions, velocities, and accelerations.
To achieve this, we employ a three-stream network architecture, where each stream consists of one encoder and one decoder.
The encoders process historical position, velocity, and acceleration data separately, capturing temporal relationships within these inputs to generate position, velocity, and acceleration features.
To enhance prediction accuracy, we propose a hierarchical feature fusion strategy across the three streams. 
This strategy integrates velocity features to assist position prediction and acceleration features to aid velocity prediction. 
Furthermore, we employ social decoders for position, velocity, and acceleration predictions. These social decoders utilize attention mechanisms to analyze neighboring features, effectively capturing interactions between pedestrians.

The predicted velocity and acceleration may not be directly correlated with the predicted position. However, they maintain strong physical consistency with historical motion. For instance, the direction of predicted velocity can be compared with the direction of historical velocity to assess alignment, while acceleration reflects changes in motion direction. This physical continuity makes velocity and acceleration effective indicators for modeling motion consistency, helping the model focus on abnormal or physically implausible position predictions.

In our framework, we model motion consistency through velocity and acceleration. Pseudo velocity and acceleration are computed from the predicted positions and treated as pseudo labels. During optimization, the model learns from these pseudo labels, which are generated directly from data rather than ground-truth annotations. Among all predicted motion states, the pair of velocity and acceleration with the highest motion consistency is selected as the best reference. We then apply a cross-entropy loss with a one-hot target to encourage the model to produce position predictions that best align with this motion pattern.

In summary, we propose a self-supervised motion consistency framework that enables the model to learn structural patterns from the data itself, without relying on ground-truth labels. We propose a three-stream network designed to capture the features of position, velocity, and acceleration. This network hierarchically integrates these features across the three streams. Experiments conducted on the ETH-UCY and SDD datasets demonstrate that the proposed method achieves state-of-the-art performance across standard trajectory prediction benchmarks. The code is available at: \url{https://github.com/YiZhou-H/Learning-Velocity-and-Acceleration}.

    
    

\section{Related Works}
\label{sec:formatting}

\textbf{Transformer-based Trajectory Prediction Model.} The core idea of transformer-based models is to utilize attention mechanisms to effectively capture social interactions among pedestrians. Several innovative methods have been proposed in this domain. Agentformer \cite{yuan2021agentformer} introduces an agent-centric attention mechanism that dynamically focuses on relevant interactions, improving trajectory prediction. MID \cite{gu2022stochastic} employs a transformer to capture temporal dependencies, with state embeddings guiding a transformer-based decoder for trajectory generation. TUTR \cite{shi2023trajectory} leverages a transformer to manage spatial-temporal dependencies, integrating post-processing techniques to enhance prediction accuracy. LAformer \cite{Liu_2024_CVPR} combines cross-attention for aligning trajectories with scene data and self-attention for refining predictions, reducing errors, and ensuring scene compliance. ENCORE-D \cite{10610614} introduces a benchmarking framework for egocentric pedestrian trajectory prediction, focusing on scenario-based evaluation to better address challenges. It also presents a novel model that integrates multimodal data through a stepwise hierarchical fusion approach. Finally, attention-aware SGTN \cite{10504962} combines graph convolutional and Transformer networks to improve stochastic trajectory prediction by effectively modeling social interactions in mixed traffic scenarios.

\textbf{Trajectory Prediction Model using Velocity. }STGAT \cite{huang2019stgat} introduces spatial-temporal graph attention networks to model interactions among pedestrians, incorporating velocity and acceleration to capture dynamic movements. Trajectory$++$ \cite{salzmann2020trajectron++} enhances this by integrating high-order dynamics, such as velocity and acceleration, alongside temporal and spatial features to improve prediction accuracy. SIT \cite{shi2022social} uses velocity to guide trajectory tree splits, predicting the most plausible future paths. MSRL \cite{wu2023multi} processes position and velocity data through temporal, spatial, and cross-correlation branches to capture pedestrian dynamics and enhance predictions. SocialCircle \cite{Wong_2024_CVPR} characterizes social interactions using velocity, distance, and direction to model the relationships between the target agent and its neighbors. Flow-Guided Markov Neural Operator (FlowMNO) \cite{10505805} highlights velocity features to enhance Trajectory$++$ and AgentFormer, improving prediction accuracy. These models collectively integrate velocity and acceleration with positional data, providing a more comprehensive understanding of pedestrian dynamics crucial for autonomous driving systems.

\section{Our Method}
\subsection{Overview}
Trajectory prediction aims to infer future trajectory based on the historical inputs. Given the historical trajectory of $i$-th pedestrian $P_i = \{(x_i^1, y_i^1), \ldots, (x_i^{\text{T}}, y_i^{\text{T}})\}$, where $i$ indexes the pedestrian, it forecasts the future movement paths $\hat{P}_i = \{(x_i^1, y_i^1), \ldots, (x_i^{\text{T}'}, y_i^{\text{T}'})\}$.
We define $\text{T}$ and $\text{T}'$ to represent the length of input and output, $N$ to represent the number of pedestrians, and the $[]$ operation such that $P_i[j]= (x_i^j, y_i^j)$ to acquire the $j$-th element from the list.

In this work, we propose to leverage velocity and acceleration information for accurate trajectory prediction. Our input contains historical position $P_{i}$, velocity, $V_{i}$ and acceleration $A_{i}$.
We calculate the $V_{i}$ and $A_{i}$ from $P_{i}$, where $V_i[j] = P_i[j+1] - P_i[j]$, and $A_i[j] = V_i[j+1] - V_i[j]$.
Our method comprises a three-stream network to predict position, velocity, and acceleration, along with a self-supervised strategy to preserve motion consistency among these three predictions. Next, we sequentially introduce the three prediction tasks and the motion consistency strategy, as illustrated in Figure 2.

\subsection{Position Prediction}
Our work leverages historical positional input to predict future trajectories. We utilize a transformer-based network for this prediction. The network comprises an encoder to capture historical temporal information and a social decoder to predict future positions by incorporating neighboring spatial information. Similar to previous approaches \cite{gu2022stochastic, Liu_2024_CVPR}, our network generates $K$ potential trajectories for each pedestrian, with the objective that at least one of these trajectories closely approximates the ground truth.

Our network employs a four-layer transformer as the encoder. Each pedestrian's position is first embedded and then processed through the transformer. The primary goal of the encoder is to capture the temporal dynamics of each pedestrian trajectory. To achieve this, we stack the trajectory of each pedestrian, and the encoder performs self-attention along the temporal dimension.
We also use a four-layer transformer to build a social decoder for $K$ potential trajectory generations. Considering the social interactions \cite{zhu2024social} between pedestrians in a scene, we reconstruct the input sequence of the decoder by obtaining the encoded features of all pedestrians within the scene.
The decoder performs self-attention along the pedestrian dimension, enabling it to capture the relationships between pedestrians. We initialize $K$ distinct query vectors to generate $K$ potential trajectories. Each query vector, combined with the reconstructed sequence, is input into the decoder to produce one potential trajectory. This process is repeated $K$ times.
The final output of each pedestrian from the positional decoder is denoted as $\hat{P}_i^{k}$, where $k$ indexes the $k$-th prediction among the $K$ potential trajectories.

Conventional methods \cite{kothari2021interpretable} typically involve manually selecting the best prediction from $K$ potential trajectories and computing the mean squared error (MSE) loss between the selected prediction and the ground truth.
In this work, we define a tolerance interval as $[P_i^{\text{G}} - \epsilon, P_i^{\text{G}} + \epsilon]$, where $\epsilon$ represents a tolerance margin and $P^{\text{G}}$ is the ground truth of position. We calculate the error between predictions and ground truth as:

\begin{equation}
\label{deqn_ex1}
\text{DIFF}{(P_i^{\text{G}},\hat{P}_i^k)} =
\begin{cases}
0 & \text{if } \hat{P}_i^k \in [P_i^{\text{G}} - \epsilon, P_i^{\text{G}} + \epsilon] \\
|\hat{P}_i^k - (P_i^{\text{G}} + \epsilon)| & \text{if } \hat{P}_i^k > P_i^{\text{G}} + \epsilon \\
|\hat{P}_i^k - (P_i^{\text{G}} - \epsilon)| & \text{if } \hat{P}_i^k < P_i^{\text{G}} - \epsilon,
\end{cases}
\end{equation}

We applied the loss to all $K$ predictions and also calculated the variance (VAR) between $K$ predictions as:
\begin{equation}
    \label{deqn_ex2}
\mathcal{L}_{\text{pos}} = \frac{1}{N}\sum_{i=1}^{N}\left\{ \alpha \sum_{k=1}^{K} \mathrm{DIFF}\left(\hat{P}_i^k, P_i^{\text{G}}\right) + \beta \text{VAR}\left(P_i^1,\dots, P_i^k\right)\right\}.
\end{equation}

\noindent The $N$ represents the number of all pedestrians; the $\alpha$ and $\beta$ are super-parameters, we empirically set to 0.2 and 0.8. Our method encourages diverse outputs through specific design elements, such as the tolerance interval in the first term, which sets the loss to zero when the prediction falls within this interval. Additionally, we include a variance loss term to prevent the predictions from becoming excessively diverse, as this could otherwise lead to a decline in performance.

\begin{figure*}[ht]
\centering
\includegraphics[width=1.0\textwidth]{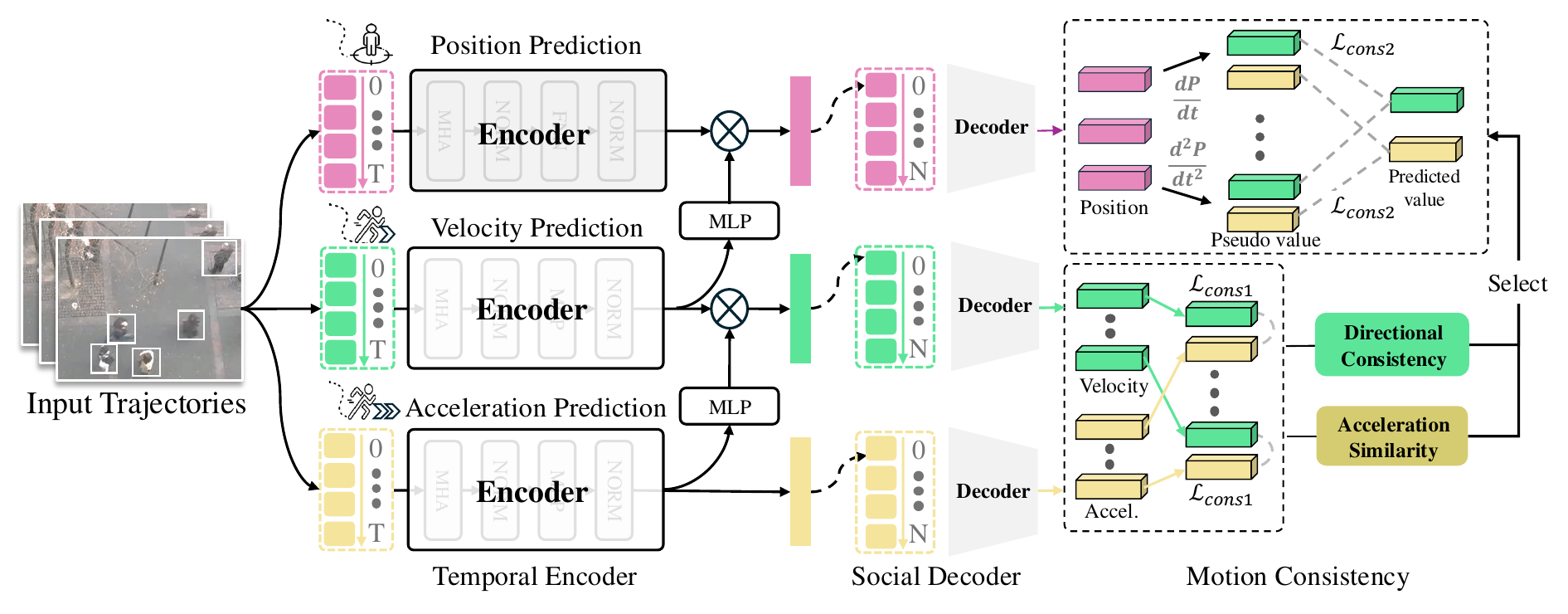} 
\caption{Our method takes position, velocity, and acceleration as inputs to a three-stream network. This network utilizes a transformer encoder to capture temporal information and outputs the features of position, velocity, and acceleration for each pedestrian. Velocity features are hierarchically injected to aid position prediction, while acceleration features are incorporated to support velocity prediction. A social decoder is then employed to predict $K$ potential trajectories based on these features. The decoder applies an attention mechanism to neighboring features, capturing interactions between pedestrians. Finally, we propose a self-supervised strategy to ensure motion consistency among position, velocity, and acceleration predictions. We group the velocity and acceleration predictions and apply $\mathcal{L}_{\text{cons1}}$ to ensure consistency within each group. Two heuristic strategies are defined to evaluate the velocity and acceleration predictions, with learnable weights assigned to these evaluations. Based on these evaluations, the network selects a velocity-acceleration group and applies $\mathcal{L}_{\text{cons2}}$ to enforce motion consistency between the position and the selected group.}
\label{Figure 2}
\end{figure*}

\subsection{Velocity and Acceleration Prediction}
We design a network to predict velocity and acceleration based on historical velocity and acceleration data. The architecture of this network mirrors that of the network used for position prediction. Temporal velocity and acceleration data for each pedestrian are input into the encoders to generate velocity and acceleration features. A social decoder is then employed to predict $K$ potential velocity and acceleration outputs. Heuristic strategies are defined to evaluate the predicted velocity and acceleration.

Global velocities are computed using the first and last frames of the historical input. To evaluate the difference between predicted velocities and global velocities, we define a directional consistency (DC) metric, where this difference is represented as angular error. Specifically, the directional consistency is formulated as:
\begin{equation}
    \label{deqn_ex3}
    DC(i, k) = 
     = \frac{v_{\text{global}} \cdot \hat{V}_i^{k}[ 1 ]}{|v_{\text{global}}| |\hat{V}_i^{k} [ 1 ]|},
\end{equation}
We use the first velocity in the prediction to compute DC, as it typically distinguishes the $K$ predictions and indicates the potential direction of the subsequent velocity predictions.

Regarding the acceleration prediction, we define a similarity to evaluate the predicted acceleration with historic acceleration. We calculate the mean value $\mu$ and variance $\sigma$ of predicted and historic acceleration and evaluate the similarity with:
\begin{equation}
    \label{deqn_ex4}
    Sim(i, k) 
     = \sqrt{ (\mu_{\text{i}} - \hat{\mu}_{i}^k)^2 + (\sigma_{\text{i}} - {\hat{\sigma}_{i}^k)}^2},
\end{equation}

We individually select the predicted velocity and acceleration from the $K$ predictions: the velocity with the largest DC and the acceleration with the smallest similarity. At this stage, the selected velocity and acceleration are not correlated. Our strategy involves independently optimizing the selected velocity and acceleration. This approach enables the optimization process to apply to distinct sets of velocity and acceleration, thereby enhancing the diversity of predictions.
We then apply the Huber Loss  \cite{huber1992robust} between the selected velocity and the ground truth and repeat the same operation for the selected acceleration to optimize the network. The resulting loss function is denoted as $\mathcal{L}_{va}$.

\subsection{Self-Supervised Motion Consistency}
We previously introduced the position, velocity, and acceleration predictions individually. In this section, we present our strategy to integrate these predictions and establish connections among them.

We first propose a feature injection strategy to fuse the features for the three predictions. The core idea is to hierarchically inject features, where velocity features are injected for position prediction and acceleration features are injected for velocity prediction. The outputs of the three encoders represent position, velocity, and acceleration features, respectively.
To facilitate this integration, we perform cross-attention between the position features and velocity features, where the velocity features serve as the query and the position features serve as the key and value. A similar operation is applied to the acceleration features for velocity prediction.

In addition to feature injection, we aim to preserve motion consistency among the three predictions during network optimization. This means that the most accurate position prediction should correspond to high-quality velocity and acceleration predictions. We first propose a consistency loss to maintain alignment between velocity and acceleration predictions.
Given the predicted velocity $\hat{V}_i^k$, we first calculate a pseudo acceleration prediction $\hat{A'}_i^k$ leveraging $\hat{V}_i^k$. We perform MSE loss between the pseudo-prediction with predicted acceleration as:

\begin{equation}
    \label{deqn_ex5}
    \mathcal{L}_{\text{cons1}} = \frac{1}{NK}\sum_{i=1}^N\sum_{k=1}^K \text{MSE}(\hat{A'}_i^k, \hat{A}_i^k),
\end{equation}
This loss implicitly groups velocity and acceleration predictions, preserving motion consistency within each group. By ensuring consistency between velocity and acceleration within the prediction set, this step eliminates potential ambiguities during model optimization.

We attempted to apply the same consistency loss between position and velocity predictions; however, it did not yield satisfactory results. To address this issue, our strategy involves first selecting the best-predicted velocity and acceleration and then preserving consistency between all position predictions and the selected pair.
Specifically, we leverage the direction consistency (Eq.\ref{deqn_ex3}) and the acceleration similarity (Eq.\ref{deqn_ex4}) to compute a score for all velocity and acceleration predictions.
{\begin{equation}
    \label{deqn_ex6}
    \begin{aligned}
         S_i^k &=  W_\alpha \cdot \text{Softmax}\left( [DC(i,1), \dots ,DC(i,k)] \right) \\
         &\quad + W_\beta \cdot \text{Softmax}\left( [Sim(i,1), \dots , Sim(i,k)] \right).
    \end{aligned}
\end{equation}}

\noindent We perform softmax between the directional consistency and the acceleration similarity of $K$ prediction for normalization, and the $Softmax()_k$ means the $k$-th output of softmax function.
$W_\alpha$ and $W_\beta$ are learnable weights.

We select one velocity prediction $\hat{V}_i^{sel}$ and one acceleration prediction $\hat{A}_i^{sel}$ with the highest score among the $K$ predictions.
Next, we compute the pseudo-velocity predictions, $\hat{V'}_i^k$ and pseudo-acceleration predictions, $\hat{A'}_i^k$ predictions for all $K$ position predictions, $\hat{P}_i^k$.
The loss function is formulated by treating $\hat{V}_i^{sel}$ and $\hat{A}_i^{sel}$ as the ground truth and enforcing the pseudo-values $\hat{V'}_i^k$ and $\hat{A'}_i^k$ to align with the ground truth. Following TUTR \cite{shi2023trajectory}, we use the cross-entropy (CE) loss to implement this function. Specifically, we compute the L2 distance between the predicted values and the ground truth. The distances are then converted into probabilities using a softmax layer, and the largest probability is maximized to $1$.
The loss function can be formulated as:

{\begin{equation}
    \label{deqn_ex7}
    \begin{aligned}
        \mathcal{L}_{\text{cons2}} = \frac{1}{2NK} \sum_{i=1}^N \sum_{k=1}^K &\Big\{ {CE}(\hat{V'}_i^k, \hat{V}_i^{sel})
        &+ {CE}(\hat{A'}_i^k, \hat{A}_i^{sel}) \Big\},
    \end{aligned}
\end{equation}}

\subsection{Implementation Details}
The loss function is summarized as:
\begin{equation}
\label{deqn_ex8}
\mathcal{L}_{\text{total}} = \mathcal{L}_{\text{pos}} + 
\mathcal{L}_{va} + \mathcal{L}_{\text{cons1}}  + \lambda \mathcal{L}_{\text{cons2}}.
\end{equation}
where $\lambda$ is the weight coefficient of the loss term to balance the importance of CE loss of velocity and acceleration.
We implement our model on Pytorch \cite{paszke1912imperative} with Adam Optimizer \cite{KingBa15} using two NVIDIA RTX 3090 24 GB GPUs. We set the learning rate to 1e-4, the epoch to 500, and the batch size to 32. 

\begin{table*}[t]
\caption{Comparison with existing state-of-the-art methods on the ETH-UCY datasets \cite{lerner2007crowds, pellegrini2009you}, bold number respresent the best performence. Notably, we campare backbone network of benchmarks to sepecify the performance gains are attributed to the introduction of backbone network of benchmarks itself. }
\label{table1}
\begin{center}
\renewcommand\arraystretch{1.1}
\tabcolsep=0.20cm
\vspace{-10pt}
\begin{tabular}{c|cccccccccc|cc}
\toprule
     \multirow{2}{*}{Methods}  & \multicolumn{2}{c|}{HOTEL $\downarrow$} & \multicolumn{2}{c|}{UNIV $\downarrow$} & \multicolumn{2}{c|}{ZARA1 $\downarrow$} & \multicolumn{2}{c|}{ZARA2 $\downarrow$} &\multicolumn{2}{c|}{ETH $\downarrow$} & \multicolumn{2}{c}{AVG $\downarrow$} \\ \cline{2-13}
        &ADE & FDE & ADE & FDE & ADE & FDE & ADE & FDE & ADE & FDE & ADE & FDE \\ 
        \midrule
\centering
PECNet \cite{mangalam2020not} &0.18&0.24&0.35&0.60&0.22&0.39&0.17&0.30& 0.54&0.87&0.29&0.48\\
AgentFormer \cite{yuan2021agentformer} & 0.14&0.22 & 0.25&0.45 & 0.18&0.30 & 0.14&0.24 & 0.45&0.75 & 0.23&0.39\\
Trajectory$++$ \cite{salzmann2020trajectron++} &0.20&0.44&0.20&0.44&0.15&0.33&0.11&0.25&0.39&0.83&0.19&0.41\\
MID \cite{gu2022stochastic}  & 0.13&0.22 & 0.22&0.45 & 0.17&0.30 & 0.13&0.27 &0.39&0.66& 0.21&0.38\\
TUTR \cite{shi2023trajectory}&0.11&0.18&0.23&0.42&0.18&0.34&0.13&0.25&0.40&0.61 &0.21&0.36\\
MemoNet \cite{xu2022remember}    & 0.11&0.17 & 0.24&0.43 & 0.18&0.32 & 0.14&0.24& 0.40&0.61 & 0.21&0.35\\
SIT \cite{shi2022social} 
&0.14&0.22&0.27&0.47&0.19&0.33&0.16&0.29 &0.39&0.62 &0.23&0.38\\
EigenTrajectory+HG \cite{kim2024higher}&0.13&0.21&0.23&0.47&0.19&0.33&0.15&0.25&0.33&0.56&0.21&0.36\\
BCDiff \cite{li2024bcdiff} &0.13&0.20&0.25&0.52&0.18&0.37&0.14&0.31&0.30&0.56&0.19&0.39\\
MSN \cite{wong2023msn}&0.11&0.17&0.28&0.48&0.22&0.36 &0.18&0.29  &\textbf{0.27}&0.41&0.21&0.35\\
MSRL \cite{wu2023multi}  &0.14&0.22&0.24&0.43&0.17&0.30&0.14&0.23&0.28&0.47&0.19&0.33\\
SocialCircle-MSN \cite{Wong_2024_CVPR}  &013&0.18&0.22&0.45&0.18&0.34&0.15&0.27&\textbf{0.27}&\textbf{0.39}&0.19&0.33\\
\hline
\textbf{Ours}&
\textbf{0.10} &\textbf{0.15} &\textbf{0.18} &\textbf{0.37} &\textbf{0.14} &\textbf{0.28} &\textbf{0.10} &\textbf{0.21} &0.29 &0.41 &\textbf{0.16} &\textbf{0.28}\\
\bottomrule
\end{tabular}
\end{center}
\end{table*}

\begin{table}[t]
\caption{Comparison with SOTA methods on SDD dataset \cite{robicquet2016learning}. Our method achieves the best performance among compared methods.}
\label{table2}
\begin{center}
\renewcommand\arraystretch{1.1}
\tabcolsep=0.20cm
\vspace{-10pt}
\begin{tabular}{c|cc}
\toprule
\multirow{2}{*}{Methods} &\multicolumn{2}{c}{SDD $\downarrow$}\\
\cline{2-3}
&ADE & FDE\\
\hline
PECNet \cite{mangalam2020not}     & 9.96 & 15.88\\
MenoNet \cite{xu2022remember}     & 9.50 & 14.78\\
MSRL \cite{wu2023multi}           & 8.22 & 13.39\\
TUTR \cite{shi2023trajectory}     & 7.76 & 12.69\\
MID \cite{gu2022stochastic}       &7.61&14.30\\
SIT \cite{shi2022social}          & 9.13 & 15.42\\
BCDiff \cite{li2024bcdiff}        & 9.05 & 14.86\\
MSRL \cite{wu2023multi} & 8.22 & 13.39\\
MSN \cite{wong2023msn}            & 7.68 & 12.39\\
EigenTrajectory+HG \cite{kim2024higher}& 7.81 & 11.09\\
SocialCircle-MSN \cite{Wong_2024_CVPR}& 7.49 & 12.12\\
\hline
\textbf{Ours}             &\textbf{7.08} & \textbf{10.47}\\
\bottomrule
\end{tabular}
\end{center}
\end{table}

\begin{table}[t]
\caption{We perform ablation study in this experiment. We remove the each component in the loss function and delete the feature injection in our network.}
\label{table3}
\begin{center}
\renewcommand\arraystretch{1.1}
\tabcolsep=0.10cm
\vspace{-10pt}
\begin{tabular}{lcccc}
\toprule
 & \multicolumn{2}{c}{\text{ETH-UCY}}&\multicolumn{2}{c}{\text{SDD}}\\ \cline{2-5}
& ADE&FDE & ADE&FDE \\
\hline
$w/o$ $\mathcal{L}_{\text{pos}}$   &0.21&0.34 & 9.41&12.72 \\

$w/o$ $\mathcal{L}_{\text{cons1}}$ & 0.25 &0.41 & 10.92 & 15.78 \\

$w/o$ $\mathcal{L}_{va}$           & 0.28 &0.47 & 12.01 & 17.61 \\

$w/o$ $\mathcal{L}_{cons2}$        & 0.39 &0.62 & 15.83 & 26.19 \\

$w/o$ Feature Injection            & 0.27 & 0.39 & 9.52 & 14.31 \\
\hline
\textbf{Ours}& \textbf{0.16}&\textbf{0.28}&\textbf{7.08}&\textbf{10.47}\\
\bottomrule
\end{tabular}
\end{center}
\end{table}

\section{Experiments}
\subsection{Datasets}
In this study, we utilize two widely used benchmark datasets for pedestrian trajectory prediction: ETH-UCY \cite{lerner2007crowds, pellegrini2009you} and SDD \cite{robicquet2016learning}. The ETH-UCY dataset comprises five sub-scenes (ETH, HOTEL, UNIV, ZARA1, and ZARA2), with trajectories recorded in meters in a world coordinate system, while the SDD dataset consists of 20 scenes, with trajectories recorded in pixels within a pixel coordinate system.

For ETH-UCY, we adopt a leave-one-out cross-validation approach \cite{shi2023representing, Wong_2024_CVPR}, training the model on four subscenes and evaluating it on the fifth. For SDD, we follow the standard train-test split defined in \cite{mangalam2020not}. Trajectory inputs are segmented into 8-second intervals, comprising 3.2 seconds of observed trajectory and 4.8 seconds of future trajectory.

\subsection{Evaluation and Metrics}
We evaluate our method using two widely accepted metrics: Average Displacement Error (ADE) and Final Displacement Error (FDE). ADE measures the average L2 distance between the predicted positions and the ground truth across all time steps, while FDE calculates the L2 distance between the predicted position and the ground truth at the final time step. To determine the ADE and FDE of our method, we follow existing state-of-the-art (SOTA) methods, set $K$ = 20 and report the performance of the trajectory that is closest to the ground truth.

\begin{figure*}[t]
\centering
\includegraphics[width=1.0\textwidth]{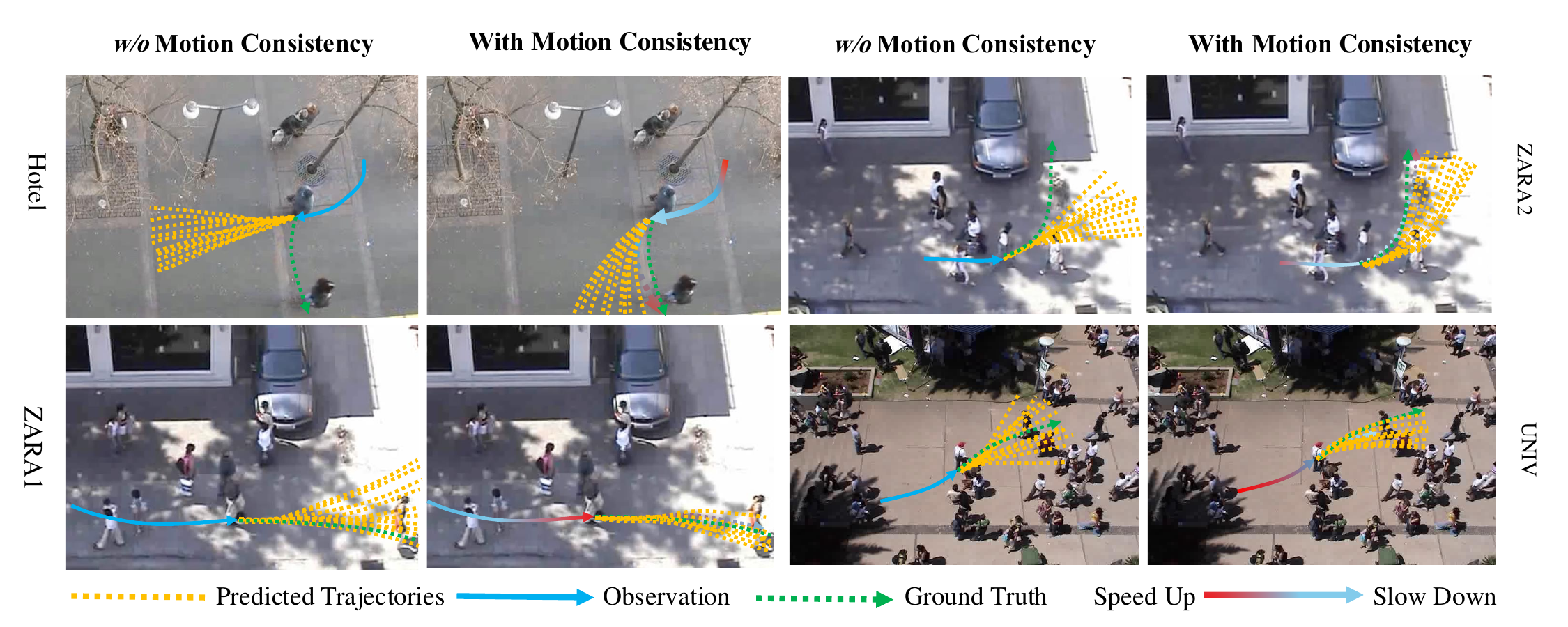} 
\caption{Visualizations of position and velocity trajectories reveal that the predicted trajectory distribution struggles to accurately account for sudden changes in a pedestrian's direction when motion consistency is not applied (\textit{w/o} motion consistency). In contrast, our method considers the pedestrian's acceleration and deceleration patterns, enabling more precise trajectory predictions aligned with the direction of the historical trajectory.}
\label{fig3}
\end{figure*}

\subsection{Comparison with SOTA Methods}
We conducted comprehensive experiments using baseline methods on the ETH-UCY and SDD datasets to evaluate the effectiveness of incorporating high-order velocities and accelerations for assessing motion consistency.

Table \ref{table1} presents the results on the ETH-UCY dataset using ADE/FDE metrics, showcasing that our method achieves SOTA performance. Specifically, our approach demonstrates a 16\% improvement in ADE and a 15\% improvement in FDE compared to SocialCircle-MSN \cite{Wong_2024_CVPR} and Multi-Stream Learning \cite{wu2023multi}. On the ZARA2 sub-dataset, our method further achieves a 10\% improvement in ADE over TUTR \cite{shi2023trajectory} and an 8\% improvement in FDE compared to MSRL \cite{wu2023multi}.

Table \ref{table2} presents the results on the SDD dataset, highlighting that our method surpasses existing SOTA approaches with a 5\% improvement in ADE compared to SocialCircle-MSN \cite{Wong_2024_CVPR} and a 6\% improvement in FDE over the second-best method, EigenTrajectory+HG \cite{kim2024higher}.

\begin{table}[t]
\caption{We replaced our module with common alternatives to assess its effectiveness. Specifically, we used the conventional MSE loss to replace $\mathcal{L}_{pos}$ and manually selected the best velocity and acceleration predictions instead of leveraging the two heuristic strategies $\mathcal{L}_{va}$. The resulting performance drop highlights the advantages of our method.}
\label{table4}
\begin{center}
\renewcommand\arraystretch{1.2}
\tabcolsep=0.35cm
\vspace{-10pt}
\begin{tabular}{ccccc}
\toprule
\multirow{2}{*}{Replacement} & \multicolumn{2}{c}{\text{ETH-UCY}}&\multicolumn{2}{c}{\text{SDD}}\\ \cline{2-5}
           & ADE&FDE & ADE&FDE \\ 
\hline

$\mathcal{L}_{pos}$ & 0.23 & 035 & 8.56 &12.71 \\
$\mathcal{L}_{va}$ & 0.26 & 0.39 & 9.88 & 14.44\\
\hline
\textbf{Ours}& \textbf{0.16}&\textbf{0.28}&\textbf{7.08}&\textbf{10.47}\\
\bottomrule
\end{tabular}
\end{center}
\end{table}

\begin{figure*}[t]
\centering
\includegraphics[width=1.0\textwidth]{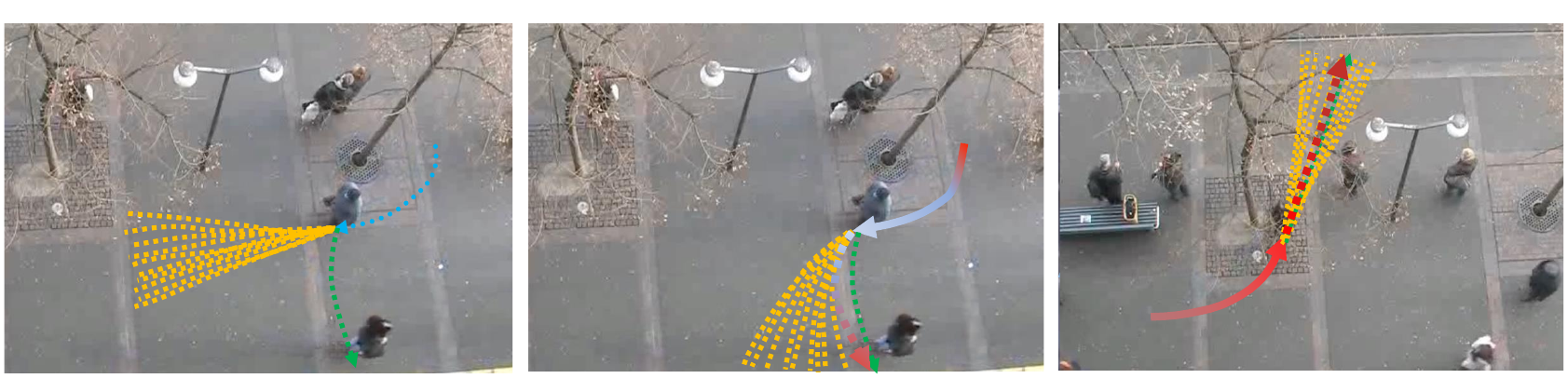} 
\caption{ The left figure shows predictions based solely on position information while the person changes movement direction. The ground truth is highlighted in \textcolor{sgreen}{green}. The middle figure shows the predictions of our method, where speed changes are visualized as \textcolor{red}{fast} and \textcolor{blue}{slow}. In this case, the person adjusts their movement direction with a slow-down signal. The right figure illustrates a scenario where the person aims to maintain their movement direction, ensuring that the speed does not decrease. This figure shows the advantage of using position, velocity and acceleration information.}
\label{fig4}
\end{figure*}

\subsection{Ablation Study}
We conducted ablation studies to evaluate the effectiveness of each component in our method. These components include the position prediction loss ($\mathcal{L}_{\text{pos}}$), the consistency loss ($\mathcal{L}_{\text{cons1}}$) for enforcing velocity and acceleration predictions, the Huber loss $(\mathcal{L}_{va})$ for $\hat{V}^{sel}$ and $\hat{A}^{sel}$, the loss for the position pseudo-values, the consistency loss for the selected velocity and acceleration based on the motion consistency strategy ($\mathcal{L}_{\text{cons2}}$) and the hierarchical feature injection strategy.
As shown in Table \ref{table3}, removing any of these components results in significant performance drops, highlighting the effectiveness of our design.
Notably, as shown in Table \ref{table4}, we conducted additional experiments to evaluate the impact of replacing the position prediction loss ($\mathcal{L}_{\text{pos}}$) ith the conventional MSE function. We also replaced the loss function $\mathcal{L}_{\text{va}}$, by bypassing the motion consistency mechanism for selecting velocity and acceleration. Instead, ground truth was used to compute the loss with $K$ predicted velocities and accelerations, and the combination yielding the minimal loss was selected as the output.

\subsection{Visualization}
To demonstrate the effectiveness of our motion consistency strategy, we visualized the predicted trajectories in typical scenarios. As illustrated in Figure \ref{fig3}, without motion consistency, the predicted trajectory distribution fails to account for sudden changes in a pedestrian's direction, leading to significant deviations from the ground truth. In contrast, when motion consistency is applied, the predicted trajectory distribution adapts more effectively to scenarios involving sudden directional changes. Additionally, for pedestrians without sudden direction changes, our method produces a more concentrated predicted trajectory distribution. This is because the predicted trajectories are matched to the selected motion pattern through pseudo-values, and the optimization process further refines the predictions around this pattern.

\begin{table}[t]
\caption{Performance of different $K$ predictions on SDD dataset.}
\label{table5}
\begin{center}
\renewcommand\arraystretch{1.1}
\tabcolsep=0.30cm
\vspace{-10pt}
\begin{tabular}{cccc}
\toprule
\multirow{2}{*}{$K$}& \multirow{2}{*}{Methods} & \multicolumn{2}{c}{\text{SDD} $\downarrow$} \\
    \cline{3-4}
    && ADE & FDE\\ 
    \hline
5      & \begin{tabular}[c]{@{}c@{}}  SIT \cite{shi2022social}\\ Y-net\cite{mangalam2021goals}\\ \textbf{Ours}
\end{tabular}  
& \begin{tabular}[c]{@{}c@{}} 12.30\\ 11.49\\\textbf{10.54}\end{tabular} & \begin{tabular}[c]{@{}c@{}} 23.17\\ 20.23\\\textbf{17.22}\end{tabular} \\
\hline
10      & \begin{tabular}[c]{@{}c@{}}  SIT \cite{shi2022social}\\Y-net\cite{mangalam2021goals}\\\textbf{Ours}
\end{tabular}  
& \begin{tabular}[c]{@{}c@{}} 10.34\\ 9.47\\\textbf{9.01}\end{tabular} & \begin{tabular}[c]{@{}c@{}} 18.50\\ 17.41\\\textbf{13.97}\end{tabular} \\
\hline
15      & \begin{tabular}[c]{@{}c@{}} SIT \cite{shi2022social}\\Y-net\cite{mangalam2021goals}\\\textbf{Ours}
\end{tabular}  
& \begin{tabular}[c]{@{}c@{}} 9.48\\ 8.44\\\textbf{7.72}\end{tabular} & \begin{tabular}[c]{@{}c@{}} 16.45\\14.17 \\\textbf{12.08}\end{tabular} \\
\bottomrule
\end{tabular}
\end{center}
\end{table}

\subsection{Case Study}
We illustrate the output in Figure \ref{fig4} to highlight the advantage of incorporating velocity and acceleration information. When a pedestrian accelerates, the likelihood of maintaining their current direction increases, while deceleration raises the probability of a directional change. In contrast, relying solely on positional information fails to capture these subtle motion variations, particularly during sudden directional changes, leading to potential inaccuracies in trajectory prediction. By integrating velocity and acceleration data, our method accurately identifies directional changes during deceleration, resulting in more precise predictions of a pedestrian's future trajectory.

\begin{table}[t]
\caption{Comparison of long-term prediction on ETH-UCY dataset, Where $\text{T}'$ refers to the predicted time steps.}
\label{table6}
\begin{center}
\renewcommand\arraystretch{1.1}
\tabcolsep=0.30cm
\vspace{-10pt}
\begin{tabular}{cccc}
\toprule
    \multirow{2}{*}{$\text{T}'$}& \multirow{2}{*}{Methods} & \multicolumn{2}{c}{\text{ETH-UCY} $\downarrow$} \\
    \cline{3-4}
    && ADE & FDE\\ 

    \hline
16         & \begin{tabular}[c]{@{}c@{}} PECNet\cite{mangalam2020not} \\ SIT \cite{shi2022social} \\\textbf{Ours}
\end{tabular}    
& \begin{tabular}[c]{@{}c@{}} 2.89\\ 0.49\\ \textbf{0.42}\end{tabular} & \begin{tabular}[c]{@{}c@{}}2.63 \\ 1.01\\ \textbf{0.87}\end{tabular} \\
\hline
20        & \begin{tabular}[c]{@{}c@{}}PECNet\cite{mangalam2020not} \\ SIT \cite{shi2022social}\\\textbf{Ours}
\end{tabular} & \begin{tabular}[c]{@{}c@{}} 3.02\\ 0.55 \\ \textbf{0.34}\end{tabular} & \begin{tabular}[c]{@{}c@{}}2.55 \\ 1.12\\ \textbf{0.91}\end{tabular} \\
\hline
24         & \begin{tabular}[c]{@{}c@{}} PECNet\cite{mangalam2020not}  \\ SIT \cite{shi2022social}\\\textbf{Ours}
\end{tabular}    & \begin{tabular}[c]{@{}c@{}} 3.16\\ 0.68\\ \textbf{0.57}\end{tabular} & \begin{tabular}[c]{@{}c@{}} 2.53\\ 1.12\\ \textbf{1.04}\end{tabular} \\ 
\bottomrule
\end{tabular}
\end{center}
\end{table}

\subsection{Additional Experiments}
\textbf{Different K.} We conducted experiments with different predicted $K$ values to evaluate the flexibility of our model. In this setup, $K$ was set to 5, 10, and 15, respectively. As shown in Table \ref{table5}, our model consistently outperforms the highest ADE/FDE results across all $K$ values.

\noindent \textbf{Long-term Prediction.} We conducted long-term prediction experiments to further evaluate the flexibility of our model. In the ETH-UCY dataset, the baseline method uses 8 time steps as observed trajectories and predicts 12 time steps for future trajectories. In this experiment, we extended the predicted time steps to 16, 20, and 24, respectively. As shown in Table \ref{table6}, our method outperforms all listed methods across the different long-term prediction horizons.

\section{Conclusion}
In this paper, we present a novel trajectory prediction framework that explicitly integrates position, velocity, and acceleration information through a unified three-stream network, designed to capture the distinct features of each component. To ensure motion consistency among the predicted position, velocity, and acceleration, we employ self-supervised learning. Furthermore, we introduce two heuristic strategies for evaluating velocity and acceleration predictions. Extensive experiments demonstrate the flexibility and accuracy of our method, with results showing that it achieves state-of-the-art performance across multiple tasks.

\section{Acknowledgement}
This work is supported in part by the Europe Eureka Intelligence to Drive | Move-Save-Win project (with funding from the UKRI Innovate UK project under Grant No. 10071278) as well as the Horizon Europe COVER project, No. 101086228 (with funding from UKRI grant EP/Y028031/1). Kezhi Wang would like to acknowledge the support in part by the Royal Society Industry Fellow scheme.

{
    \small
    \bibliographystyle{ieeenat_fullname}
    \bibliography{ref}

\begin{thebibliography}{36}
\providecommand{\natexlab}[1]{#1}
\providecommand{\url}[1]{\texttt{#1}}
\expandafter\ifx\csname urlstyle\endcsname\relax
  \providecommand{\doi}[1]{doi: #1}\else
  \providecommand{\doi}{doi: \begingroup \urlstyle{rm}\Url}\fi

\bibitem[Bae and Jeon(2023)]{bae2023set}
Inhwan Bae and Hae-Gon Jeon.
\newblock A set of control points conditioned pedestrian trajectory prediction.
\newblock In \emph{Proceedings of the AAAI Conference on Artificial Intelligence}, pages 6155--6165, 2023.

\bibitem[Bae et~al.(2023)Bae, Oh, and Jeon]{bae2023eigentrajectory}
Inhwan Bae, Jean Oh, and Hae-Gon Jeon.
\newblock Eigentrajectory: Low-rank descriptors for multi-modal trajectory forecasting.
\newblock In \emph{Proceedings of the IEEE/CVF International Conference on Computer Vision}, pages 10017--10029, 2023.

\bibitem[Cheng et~al.(2024)Cheng, Zhu, Wang, Hao, Liu, Cheng, Wang, and Chang]{Cheng_2024_CVPR}
Yihua Cheng, Yaning Zhu, Zongji Wang, Hongquan Hao, Yongwei Liu, Shiqing Cheng, Xi Wang, and Hyung~Jin Chang.
\newblock What do you see in vehicle? comprehensive vision solution for in-vehicle gaze estimation.
\newblock In \emph{Proceedings of the IEEE/CVF Conference on Computer Vision and Pattern Recognition (CVPR)}, pages 1556--1565, 2024.

\bibitem[Gu et~al.(2022)Gu, Chen, Li, Lin, Rao, Zhou, and Lu]{gu2022stochastic}
Tianpei Gu, Guangyi Chen, Junlong Li, Chunze Lin, Yongming Rao, Jie Zhou, and Jiwen Lu.
\newblock Stochastic trajectory prediction via motion indeterminacy diffusion.
\newblock In \emph{Proceedings of the IEEE/CVF Conference on Computer Vision and Pattern Recognition}, pages 17113--17122, 2022.

\bibitem[Huang et~al.(2019)Huang, Bi, Li, Mao, and Wang]{huang2019stgat}
Yingfan Huang, Huikun Bi, Zhaoxin Li, Tianlu Mao, and Zhaoqi Wang.
\newblock Stgat: Modeling spatial-temporal interactions for human trajectory prediction.
\newblock In \emph{Proceedings of the IEEE/CVF international conference on computer vision}, pages 6272--6281, 2019.

\bibitem[Huang et~al.(2025)Huang, Cheng, and Wang]{huang2025trajectory}
Yizhou Huang, Yihua Cheng, and Kezhi Wang.
\newblock Trajectory mamba: Efficient attention-mamba forecasting model based on selective ssm.
\newblock \emph{arXiv preprint arXiv:2503.10898}, 2025.

\bibitem[Huber(1992)]{huber1992robust}
Peter~J Huber.
\newblock Robust estimation of a location parameter.
\newblock In \emph{Breakthroughs in statistics: Methodology and distribution}, pages 492--518. Springer, 1992.

\bibitem[Jafari and Liu(2024)]{jafari2024pedestrians}
Alireza Jafari and Yen-Chen Liu.
\newblock Pedestrians' safety using projected time-to-collision to electric scooters.
\newblock \emph{Nature communications}, 15\penalty0 (1):\penalty0 5701, 2024.

\bibitem[Kim et~al.(2024)Kim, Chi, Lim, Ramani, Kim, and Kim]{kim2024higher}
Sungjune Kim, Hyung-gun Chi, Hyerin Lim, Karthik Ramani, Jinkyu Kim, and Sangpil Kim.
\newblock Higher-order relational reasoning for pedestrian trajectory prediction.
\newblock In \emph{Proceedings of the IEEE/CVF Conference on Computer Vision and Pattern Recognition}, pages 15251--15260, 2024.

\bibitem[Kingma and Ba(2015)]{KingBa15}
Diederik Kingma and Jimmy Ba.
\newblock Adam: A method for stochastic optimization.
\newblock In \emph{International Conference on Learning Representations (ICLR)}, San Diega, CA, USA, 2015.

\bibitem[Kothari et~al.(2021)Kothari, Sifringer, and Alahi]{kothari2021interpretable}
Parth Kothari, Brian Sifringer, and Alexandre Alahi.
\newblock Interpretable social anchors for human trajectory forecasting in crowds.
\newblock In \emph{Proceedings of the IEEE/CVF Conference on Computer Vision and Pattern Recognition}, pages 15556--15566, 2021.

\bibitem[Lerner et~al.(2007)Lerner, Chrysanthou, and Lischinski]{lerner2007crowds}
Alon Lerner, Yiorgos Chrysanthou, and Dani Lischinski.
\newblock Crowds by example.
\newblock In \emph{Computer graphics forum}, pages 655--664. Wiley Online Library, 2007.

\bibitem[Li et~al.(2024{\natexlab{a}})Li, Li, Ren, Chen, Yuan, and Wang]{li2024bcdiff}
Rongqing Li, Changsheng Li, Dongchun Ren, Guangyi Chen, Ye Yuan, and Guoren Wang.
\newblock Bcdiff: Bidirectional consistent diffusion for instantaneous trajectory prediction.
\newblock \emph{Advances in Neural Information Processing Systems}, 36, 2024{\natexlab{a}}.

\bibitem[Li et~al.(2024{\natexlab{b}})Li, Gong, Chen, Huang, and Zhong]{li2024integrated}
Xingyu Li, Xinle Gong, Ye-Hwa Chen, Jin Huang, and Zhihua Zhong.
\newblock Integrated path planning-control design for autonomous vehicles in intelligent transportation systems: A neural-activation approach.
\newblock \emph{IEEE Transactions on Intelligent Transportation Systems}, 2024{\natexlab{b}}.

\bibitem[Liu et~al.(2024{\natexlab{a}})Liu, Cheng, Chen, Broszio, Li, Zhao, Sester, and Yang]{Liu_2024_CVPR}
Mengmeng Liu, Hao Cheng, Lin Chen, Hellward Broszio, Jiangtao Li, Runjiang Zhao, Monika Sester, and Michael~Ying Yang.
\newblock Laformer: Trajectory prediction for autonomous driving with lane-aware scene constraints.
\newblock In \emph{Proceedings of the IEEE/CVF Conference on Computer Vision and Pattern Recognition (CVPR) Workshops}, pages 2039--2049, 2024{\natexlab{a}}.

\bibitem[Liu et~al.(2024{\natexlab{b}})Liu, Li, Wang, Sammut, and Yao]{10504962}
Yao Liu, Binghao Li, Xianzhi Wang, Claude Sammut, and Lina Yao.
\newblock Attention-aware social graph transformer networks for stochastic trajectory prediction.
\newblock \emph{IEEE Transactions on Knowledge and Data Engineering}, pages 1--14, 2024{\natexlab{b}}.

\bibitem[Mangalam et~al.(2020)Mangalam, Girase, Agarwal, Lee, Adeli, Malik, and Gaidon]{mangalam2020not}
Karttikeya Mangalam, Harshayu Girase, Shreyas Agarwal, Kuan-Hui Lee, Ehsan Adeli, Jitendra Malik, and Adrien Gaidon.
\newblock It is not the journey but the destination: Endpoint conditioned trajectory prediction.
\newblock In \emph{Computer Vision--ECCV 2020: 16th European Conference, Glasgow, UK, August 23--28, 2020, Proceedings, Part II 16}, pages 759--776. Springer, 2020.

\bibitem[Mangalam et~al.(2021)Mangalam, An, Girase, and Malik]{mangalam2021goals}
Karttikeya Mangalam, Yang An, Harshayu Girase, and Jitendra Malik.
\newblock From goals, waypoints \& paths to long term human trajectory forecasting.
\newblock In \emph{Proceedings of the IEEE/CVF International Conference on Computer Vision}, pages 15233--15242, 2021.

\bibitem[Paszke et~al.(1912)Paszke, Gross, Massa, Lerer, Bradbury, Chanan, Killeen, Lin, Gimelshein, Antiga, et~al.]{paszke1912imperative}
A Paszke, S Gross, F Massa, A Lerer, JP Bradbury, G Chanan, T Killeen, Z Lin, N Gimelshein, L Antiga, et~al.
\newblock An imperative style, high-performance deep learning library.
\newblock \emph{Adv. Neural Inf. Process. Syst}, 32:\penalty0 8026, 1912.

\bibitem[Pellegrini et~al.(2009)Pellegrini, Ess, Schindler, and Van~Gool]{pellegrini2009you}
Stefano Pellegrini, Andreas Ess, Konrad Schindler, and Luc Van~Gool.
\newblock You'll never walk alone: Modeling social behavior for multi-target tracking.
\newblock In \emph{2009 IEEE 12th international conference on computer vision}, pages 261--268. IEEE, 2009.

\bibitem[Rasouli(2024)]{10610614}
Amir Rasouli.
\newblock A novel benchmarking paradigm and a scale- and motion-aware model for egocentric pedestrian trajectory prediction.
\newblock In \emph{2024 IEEE International Conference on Robotics and Automation (ICRA)}, pages 5630--5636, 2024.

\bibitem[Robicquet et~al.(2016)Robicquet, Sadeghian, Alahi, and Savarese]{robicquet2016learning}
Alexandre Robicquet, Amir Sadeghian, Alexandre Alahi, and Silvio Savarese.
\newblock Learning social etiquette: Human trajectory understanding in crowded scenes.
\newblock In \emph{Computer Vision--ECCV 2016: 14th European Conference, Amsterdam, The Netherlands, October 11-14, 2016, Proceedings, Part VIII 14}, pages 549--565. Springer, 2016.

\bibitem[Salzmann et~al.(2020)Salzmann, Ivanovic, Chakravarty, and Pavone]{salzmann2020trajectron++}
Tim Salzmann, Boris Ivanovic, Punarjay Chakravarty, and Marco Pavone.
\newblock Trajectron++: Dynamically-feasible trajectory forecasting with heterogeneous data.
\newblock In \emph{Computer Vision--ECCV 2020: 16th European Conference, Glasgow, UK, August 23--28, 2020, Proceedings, Part XVIII 16}, pages 683--700. Springer, 2020.

\bibitem[Sarker(2021)]{sarker2021machine}
Iqbal~H Sarker.
\newblock Machine learning: Algorithms, real-world applications and research directions.
\newblock \emph{SN computer science}, 2\penalty0 (3):\penalty0 160, 2021.

\bibitem[Shi et~al.(2022)Shi, Wang, Long, Zhou, Zheng, Zheng, and Hua]{shi2022social}
Liushuai Shi, Le Wang, Chengjiang Long, Sanping Zhou, Fang Zheng, Nanning Zheng, and Gang Hua.
\newblock Social interpretable tree for pedestrian trajectory prediction.
\newblock In \emph{Proceedings of the AAAI Conference on Artificial Intelligence}, pages 2235--2243, 2022.

\bibitem[Shi et~al.(2023{\natexlab{a}})Shi, Wang, Long, Zhou, Tang, Zheng, and Hua]{shi2023representing}
Liushuai Shi, Le Wang, Chengjiang Long, Sanping Zhou, Wei Tang, Nanning Zheng, and Gang Hua.
\newblock Representing multimodal behaviors with mean location for pedestrian trajectory prediction.
\newblock \emph{IEEE transactions on pattern analysis and machine intelligence}, 2023{\natexlab{a}}.

\bibitem[Shi et~al.(2023{\natexlab{b}})Shi, Wang, Zhou, and Hua]{shi2023trajectory}
Liushuai Shi, Le Wang, Sanping Zhou, and Gang Hua.
\newblock Trajectory unified transformer for pedestrian trajectory prediction.
\newblock In \emph{Proceedings of the IEEE/CVF International Conference on Computer Vision}, pages 9675--9684, 2023{\natexlab{b}}.

\bibitem[Thumm et~al.(2024)Thumm, Trost, and Althoff]{thumm2024human}
Jakob Thumm, Felix Trost, and Matthias Althoff.
\newblock Human-robot gym: Benchmarking reinforcement learning in human-robot collaboration.
\newblock In \emph{2024 IEEE International Conference on Robotics and Automation (ICRA)}, pages 7405--7411. IEEE, 2024.

\bibitem[Uhlemann et~al.(2024)Uhlemann, Fent, and Lienkamp]{10505805}
Nico Uhlemann, Felix Fent, and Markus Lienkamp.
\newblock Evaluating pedestrian trajectory prediction methods with respect to autonomous driving.
\newblock \emph{IEEE Transactions on Intelligent Transportation Systems}, pages 1--10, 2024.

\bibitem[Wong et~al.(2023)Wong, Xia, Peng, Yuan, and You]{wong2023msn}
Conghao Wong, Beihao Xia, Qinmu Peng, Wei Yuan, and Xinge You.
\newblock Msn: multi-style network for trajectory prediction.
\newblock \emph{IEEE Transactions on Intelligent Transportation Systems}, 24\penalty0 (9):\penalty0 9751--9766, 2023.

\bibitem[Wong et~al.(2024)Wong, Xia, Zou, Wang, and You]{Wong_2024_CVPR}
Conghao Wong, Beihao Xia, Ziqian Zou, Yulong Wang, and Xinge You.
\newblock Socialcircle: Learning the angle-based social interaction representation for pedestrian trajectory prediction.
\newblock In \emph{Proceedings of the IEEE/CVF Conference on Computer Vision and Pattern Recognition (CVPR)}, pages 19005--19015, 2024.

\bibitem[Wu et~al.(2023)Wu, Wang, Zhou, Duan, Hua, and Tang]{wu2023multi}
Yuxuan Wu, Le Wang, Sanping Zhou, Jinghai Duan, Gang Hua, and Wei Tang.
\newblock Multi-stream representation learning for pedestrian trajectory prediction.
\newblock In \emph{Proceedings of the AAAI Conference on Artificial Intelligence}, pages 2875--2882, 2023.

\bibitem[Xu et~al.(2022)Xu, Mao, Zhang, and Chen]{xu2022remember}
Chenxin Xu, Weibo Mao, Wenjun Zhang, and Siheng Chen.
\newblock Remember intentions: Retrospective-memory-based trajectory prediction.
\newblock In \emph{Proceedings of the IEEE/CVF Conference on Computer Vision and Pattern Recognition}, pages 6488--6497, 2022.

\bibitem[Yuan et~al.(2021)Yuan, Weng, Ou, and Kitani]{yuan2021agentformer}
Ye Yuan, Xinshuo Weng, Yanglan Ou, and Kris~M Kitani.
\newblock Agentformer: Agent-aware transformers for socio-temporal multi-agent forecasting.
\newblock In \emph{Proceedings of the IEEE/CVF International Conference on Computer Vision}, pages 9813--9823, 2021.

\bibitem[Yue et~al.(2022)Yue, Manocha, and Wang]{yue2022human}
Jiangbei Yue, Dinesh Manocha, and He Wang.
\newblock Human trajectory prediction via neural social physics.
\newblock In \emph{European conference on computer vision}, pages 376--394. Springer, 2022.

\bibitem[Zhu et~al.(2024)Zhu, Qin, Lou, Ye, Ma, Ci, and Wang]{zhu2024social}
Wentao Zhu, Jason Qin, Yuke Lou, Hang Ye, Xiaoxuan Ma, Hai Ci, and Yizhou Wang.
\newblock Social motion prediction with cognitive hierarchies.
\newblock \emph{Advances in Neural Information Processing Systems}, 36, 2024.

\end{thebibliography}
}

\end{document}